\documentclass{article}

\usepackage{PRIMEarxiv}

\usepackage[utf8]{inputenc} 
\usepackage[T1]{fontenc}    
\usepackage{hyperref}       
\usepackage{url}            
\usepackage{booktabs}       
\usepackage{amsfonts}       
\usepackage{nicefrac}       
\usepackage{microtype}      
\usepackage{lipsum}
\usepackage{fancyhdr}       
\usepackage{graphicx}       
\graphicspath{{media/}}     
\usepackage{algorithm}

\usepackage{algpseudocode, amsmath} 

\title{The Influences of Color and Shape Features \\ in Visual Contrastive Learning}

\author{
  Xiaoqi Zhuang\\
  \texttt{xiaoqizhuang@outlook.com}
}

\begin{document}
\maketitle

\begin{abstract}
In the field of visual representation learning, performance of contrastive learning has been catching up with the supervised method which is commonly a classification convolutional neural network. However, most of the research work focuses on improving the accuracy of downstream tasks such as image classification and object detection. For visual contrastive learning, the influences of individual image features (e.g., color and shape) to model performance remain ambiguous. 

This paper investigates such influences by designing various ablation experiments, the results of which are evaluated by specifically designed metrics. While these metrics are not invented by us, we first use them in the field of representation evaluation. Specifically, we assess the contribution of two primary image features (i.e., color and shape) in a quantitative way. Experimental results show that compared with supervised representations, contrastive representations tend to cluster with objects of similar color in the representation space, and contain less shape information than supervised representations. Finally, we discuss that the current data augmentation is responsible for these results. We believe that exploring an unsupervised augmentation method that can strongly change the shape information of objects is an effective research direction that can improve unsupervised visual representations.
\end{abstract}


\section{Introduction}
\label{sec:intro}
Unsupervised visual representation learning, specifically contrastive learning, has seen rapid development in recent years \cite{he2019moco, chen2020simple, Chen_2021_CVPR}. These methods aim to train a deep neural network encoder as a feature extractor by minimizing the distance of similar objects and maximizing the distance of dissimilar ones. Due to the limit of unsupervised learning, the ground truth of each image is only itself. Therefore, the mainstream methods usually apply different types of pretext tasks. \cite{wu2018unsupervised} proposed a pretext task called instance discrimination, which applies different image transformations on each image to get a pair of positive samples and define all other images as negative samples. Previous works proposed 3 key components to make such training framework succeed: deep neural networks based image transformation \cite{chen2020simple}, a large size of negative samples, and feature consistency\cite{he2019moco, chen2020simple}. \cite{chen2020simple} showed that adding color distortion and cropping into the transformation pipeline significantly improve accuracy, which is because simple image augmentation such as rotation and flipping could not largely change the distribution. Large negative samples, which is because the positive samples are only the two augmentations from a single image, the contrasting learning loss function \cite{aaron2018cpc} always has to use thousands of negative samples to project images into the representation space. Feature consistency, \cite{he2019moco, chen2020simple} proposed that the representation features should be compared into the same representation space to avoid the biased contrastive loss from the updated encoders after backpropagation. These trivial works are the basis for contrastive learning, which achieved around 74\% accuracy in Image-Net datasets. 

Representation learning has always used improving the accuracy of downstream tasks as an indicator to improve the backbone of the model. However, we believe that the accuracy of a single downstream task such as image classification is not sufficient for the evaluation of representation quality. Therefore, We developed metrics for representation models to explicitly quantify visual representations by evaluating their color and shape information. At the same time, supervised methods have always been considered the upper limit of unsupervised methods. We also want to find out the gap between them, thus to clarify the improvement ideas of unsupervised methods. Therefore, we used ResNet-18 \cite{He2015} to train three unsupervised models and one supervised model for experiments.

In this paper, we did these experiments. Firstly, we try to find the gap between supervised and unsupervised models. We get wrong predictions from unsupervised models with image classification tasks. We found that nearly half of them were the same predictions for three unsupervised models. These images can be considered common faults of unsupervised methods, and the parts of these images where the supervised model can make correct predictions are considered as gaps. Secondly, in terms of color information, we found that these images and their false predictions are strongly correlated with color. We use the earth mover's distance \cite{Rubner:2000} metric to demonstrate that unsupervised representations are color biased within the representation space. For shape information, we generated a CIFAR-10 \cite{cifar10} silhouette dataset by a pretrained segmentation model. We verified that the accuracy of the unsupervised model on this dataset is also lower than that of the supervised model. We recomputed the accuracy of the image classification task with color and shape-distorted images, and the results show that unsupervised models are more robust than supervised models. Finally, we conclude that there is no distribution change method strongly related to the image shape among the data augmentation methods, which is the main reason for the gap with the supervised model.

\section{Related Work}
\subsection*{Unsupervised contrastive visual learning}
Due to the limitation of an unsupervised task, the ground truth of an image is just itself, leading that it being hard to design the learning progress. \cite{wu2018unsupervised} proposed a pretext task called Instance Discrimination to annotate images by taking them as positive and negative samples. The core concept is comparing the similarity between two images that are transformed by a single image. Therefore, we can train the model by reducing the distance of generated image pairs and extending the distance of other images. Such a learning framework requires two important keys \cite{chen2020simple}: large negative samples, and feature consistency. \cite{he2019moco, chen2020mocov2} proposed a first-in-first-out queue to store large negative samples, and keep feature consistency by updating a momentum encoder slowly. \cite{chen2020simple} achieved both keys by training the model within a large batch size, which is straightforward but hard to reproduce . \cite{chen2020simple} proved that adding color jitter as well as cropping into image transformation will largely improve the model. \cite{chen2020simple} added a linear layer called "Projection Layer" after the backbone model, which can help raise the accuracy by nearly 7\%. Such modifications and the training details in \cite{wu2018unsupervised} have been the fixed set of contrastive learning. \cite{grill2020bootstrap, Chen_2021_CVPR} proposed a contrastive learning framework without negative samples. They prove that only comparing the similarity of two positive images can achieve nearly the same accuracy as the state-of-the-art models without large batch size as well as momentum encoders.   

\subsection*{Improvements on unsupervised representation learning}
A lot of work has been done to improve the basic methods mentioned above. \cite{huynh2022fnc, NEURIPS2020_23af4b45, robinson2020hard} want to get high-quality positive and negative sample pairs for contrastive learning. They selected positive and negative samples instead of random selection by using similarity measures during training. \cite{peng2022crafting} used the heat map of feature representations to obtain an approximate bounding box to execute an accurate "CROP" operation. \cite{Xiao2021, dangovski2021equivariant} suggests that inappropriate data augmentations will harm image information. They use enumeration to select the right data augmentations for each image for contrastive training. \cite{caron2020unsupervised, guo2022hcsc, PCL} argue that instance discrimination can only provide instance-level representation information. Their methods can provide high-level semantic information for unsupervised learning by adding clustering methods.

\subsection*{Experiments of visual representation}
Researchers have done many experiments on representation bias. \cite{Robert2019} used pure texture and shape image datasets to illustrate that the CNN model trained on Image-Net is texture biased. \cite{Cole2022} explores the effects of contrastive learning in terms of data volume, data domain, data quality, and task granularity. \cite{Ge2022} designed a specific representation learning model for shape, color, and texture, which help them analysis the specific contribution of each attributes when the model is referenced.

In this paper, our work focuses on analyzing color and shape information between representations learned by contrastive and supervised methods. We then look for potential gaps by comparing them with supervised representations. After that, we believe that such gaps will instruct future work to improve unsupervised models.

\section{Methods}
\subsection{Dataset}
We perform experiments on a popular small image dataset: CIFAR-10 \cite{cifar10}, which consists of 60000 32x32 color images in 10 classes, with 6000 images per class. We know that the experiment results from such a small dataset may be biased. However, all these methods \cite{he2019moco, chen2020simple, Chen_2021_CVPR} show the same problems in this dataset. We believe these results are valuable to discuss.

\subsection{Training details}
\subsubsection*{Unsupervised backbone}
We trained three mainstream unsupervised models based on the ResNet-18 \cite{He2015} backbone, which are: MoCo V2 \cite{chen2020mocov2}, SimCLR V2 \cite{chen2020mocov2}, and SimSiam \cite{Chen_2021_CVPR}. We use the linear evaluation method as the metric, that is, training these unsupervised methods on the trainset in 800 epochs firstly, extracting and freezing the backbone encoder, adding an extra MLP to train a classifier on the trainset in 200 epochs, and then evaluating the accuracy of the classifier on the test set. Due to the low resolution of images in CIFAR-10, we modified the Standard ResNet-18 backbone. We delete "Gaussian blur" in the transformation compose, replace the $7 \times 7$ convolution kernel as the $3 \times 3$ kernel and delete the "stride" and "MaxPool2d". The hyperparameters of the temperature $\tau$, learning rate and scheduler are the same as \cite{wu2018unsupervised}.

\subsubsection*{Supervised backbone}
We also trained a supervised model based on ResNet-18 as the object of comparison. The model structure is largely the same as the above-unsupervised models. We add a classification layer and used labels to train the model. 

\subsubsection*{Linear Evaluation Accuracy}
The evaluation results are shown in Table \ref{tab: accuracy}. In terms of linear evaluation, three contrastive models achieved around 80\% accuracy after 200 epochs and 88\% after 800 epochs. By contrast, the accuracy of the supervised ResNet-18 is 95.18\% within only 200 epochs. 

\begin{table}
  \centering
  \begin{tabular}{c c c c c}
    \toprule
     & MoCo & SimCLR & SimSiam & Supervised \\ 
    \midrule
    200 eps & 81.86\% & 83.96\% & 80.41\% & 95.18\% \\  
    800 eps  & 88.14\% & 87.78\% & 88.43\% &  \\ 
    improve  & 6.28\% & 3.82\% & 8.02\% & \\ 
    \bottomrule
  \end{tabular}
  \caption{Accuracy on different contrastive learning backbone. All contrastive methods do not have major differences on accuracy. It also shows that the accuracy of the model increases slowly after 200 epochs, which increases by less than 10\% after training 600 epochs.}
  \label{tab: accuracy}
\end{table}

\section{Experiments}
We now describe our experiments in the following order. Firstly, we found the gap between contrastive models and the supervised model, and then proposed our assumptions based on such a gap. Secondly, we implemented specific experiments to verify these assumptions. Finally, we discussed the possible reasons for such issues on contrastive representation.

\subsection{Gap Definition}
Table \ref{tab: accuracy} shows that the accuracy difference between the three backbones in the CIFAR-10 test set is less than 1\%, which arouses our interest in whether they do the same predictions. The results show that the intersection between false predictions has 581 images (nearly 49\%), the intersection between false predictions with the same predictions has 384 images (nearly 32\%). This indicates that although these backbone models are trained in different methods, the representation features are similar in some areas as long as they are trained by instance discrimination, which also means that they do have the same problems. 

However, the supervised model can make 265 right predictions in these 384 images. Due to the low resolution and wrong labeling reasons in the CIFAR-10 dataset, we can take the supervised model as the upper limit of the classification performance. Therefore, such 265 images can be regarded as the knowledge gap between the supervised model and unsupervised models. We will call these images GAP-265 in the following content.

\begin{figure}[t]
  \centering
    \includegraphics[scale=0.6]{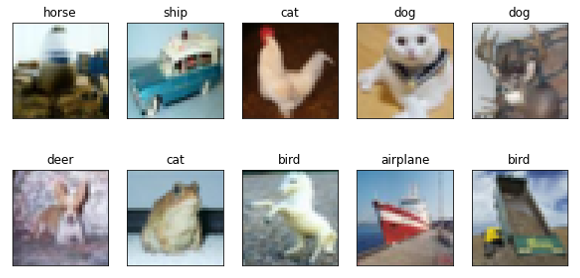}
   \caption{The same wrong predictions by contrastive methods. These images are easy to be identified for both human beings and the supervised model. However, all models based on contrastive learning make wrong predictions, which these predictions, the text over the image, are probably made by the color information. For example, the plane on the top-left corner is predicted as a horse, where the only same semantic of these two objects is the color: brown.}
   \label{fig: fps}
\end{figure}

Figure \ref{fig: fps} is shown as a subset of these images. These images are easy to be identified for both human beings and supervised model. However, all models based on contrastive learning make wrong predictions, and these predictions are likely to be made based on their color information. For example, the plane on the top-left corner is predicted as a horse, where the only same semantic of these two objects is the color: brown. 

Such phenomenon inspires us to propose two assumptions:
\begin{itemize}
\item[$\bullet$] Contrastive representation space tends to make images with similar colors closer. 
\item[$\bullet$] Contrastive features contain less shape information than supervised features.
\end{itemize}

\subsection{Feature Influence}
\begin{figure}[t]
  \centering
    \includegraphics[scale=0.3]{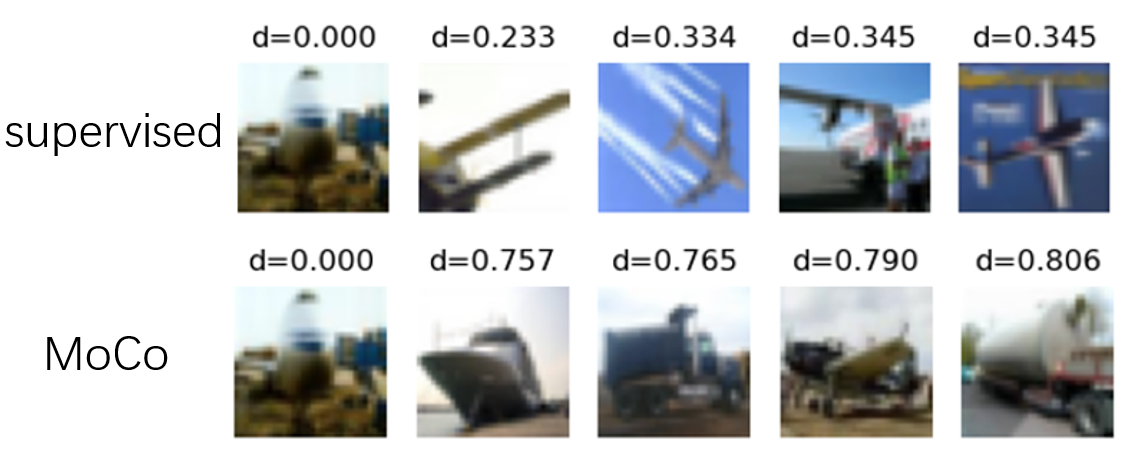}
   \caption{Two examples of the false prediction and their top-4 nearest neighbours. The target image are not in the right regions, and they are in the region which their color are similar.}
   \label{fig: knn_example}
\end{figure}

We use Linear Evaluation to compare different models, and thus we should prove that our hypothesis should not be affected by the extra linear layers. In other words, features embedded by the contrastive encoder are the reason for the wrong predictions in GAP-265. Therefore, we perform k-nearest neighbor (kNN) classification on GAP-265 to eliminate such concerns. Table \ref{tab: kNN} shows that kNN-4 accuracy of models by unsupervised methods are also largely lower than the one by the supervised method. Figure \ref{fig: knn_example} shows an example of different nearest neighbors for the brown plane between MoCo and the supervised model. The nearest neighbors in MoCo are other categories, while the only similar field is color. By contrast, all nearest neighbors in the supervised model are the same categories but different colors.

\begin{table}
  \centering
  \begin{tabular}{c c c c}
    \toprule
    MoCo & SimCLR & SimSiam & Supervised \\ 
    \midrule
    56.51\% & 62.00\% & 60.16\% & 77.89\% \\
    \bottomrule
  \end{tabular}
  \caption{kNN-4 Accuracy on GAP-265 between MoCo, SimCLR, SimSiam and supervised ResNet-18}
  \label{tab: kNN}
\end{table}

\subsection{Color Biased Detection}
\label{sec: color biased}
To prove our first assumption, contrastive representation space tends to make images with similar colors closer, we use Earth Mover's Distance(EMD) \cite{Rubner:2000}, which is also called Wasserstein Distance, as a metric to evaluate the color similarity between the query image and its top-4 nearest neighbors. Algorithm \ref{emd} illustrates the pseudo-code that how we use EMD to evaluate image similarity on color dimension. First, we convert the BGR image into its HSV form that conforms to human vision. Second, we calculate the color histogram of the image in the hue dimension, and convert it into the signature form\cite{Rubner:2000} which includes index information. Finally, We calculate EMD between the object image and the signatures of its nearest four neighbors to compare whether the image is close to the image of similar color in the representation space. More specifically, the smaller the EMD value between the query and its neighbours, the more similar they are in terms of color. Table \ref{tab: color biased} shows the mean value of EMD in GAP-265 between all backbones. It is clear that the metrics of backbones learned by contrastive methods are near and smaller than the metric of the model by the supervised method. Figure \ref{fig: knn_example} is also a clear example that the same query will have different nearest neighbours in different models, while contrastive features are clustered by color and supervised features are clustered by their labels.

\begin{algorithm}
	\renewcommand{\algorithmicrequire}{\textbf{Input:}}
	\renewcommand{\algorithmicensure}{\textbf{Output:}}
	\caption{Earth Mover's Distance on Image Color Similarity}
    \label{emd}
    \algorithmicrequire a test image $t_i$ , model $M$, the overall test set $test$, neighbor number $k$ \\
    \algorithmicensure the mean EMD between $t_i$ and its nearest K neighbors
	\begin{algorithmic}[1]
        \Function {$image2signature$} {image}
            \State bgr = CV2.imread(image)
            \State hsv = CV2.cvtColor(bgr) \Comment{convert BGR to HSV}
            \State histHUE = CV2.calcHist(hsv, channel=0) \Comment{only calculate histogram on HUE}
            \State sign = hist2signature(histHUE) \Comment{convert histogram to signature}
            \State \Return sign
        \EndFunction
        \State $dist_i$ = [] \Comment{create an empty list to store EMD values}
        \State $f_{i}, f_{all}$ = $M(t_i), M(test)$ \Comment{extract features by model}
        \State $neighbors_{i}$ = nearestNeighbors($f_{i}$, $f_{all}$, $k$) \Comment{get the nearest K neighbors of $t_i$}
        \State $sign_{i}$ = image2signature($t_i$)
        \For {$n \in neighbors_{i}$} 
            \State $sign_n$ = image2signature(n)
            \State $emd_n$ = cv2.EMD($sign_i$, $sign_n$, L2)
            \State dist.append($emd_n$)
        \EndFor
	\end{algorithmic} 
    \algorithmicensure mean($dist_i$)
\end{algorithm}

\begin{table}
  \centering
  \begin{tabular}{c c c c c}
    \toprule
    Model & MoCo & SimCLR & SimSiam & Supervised\\ 
    \midrule
    EMD & 26.97  & 28.58  & 28.85 & 33.20 \\
    \bottomrule
  \end{tabular}
  \caption{The Earth Mover's Distance in GAP-265 between contrasitve and supervised models.}
  \label{tab: color biased}
\end{table}

\subsection{Shape Biased Detection}
\label{sec: Shape biased}
\begin{figure}[t]
  \centering
    \includegraphics[scale=0.3]{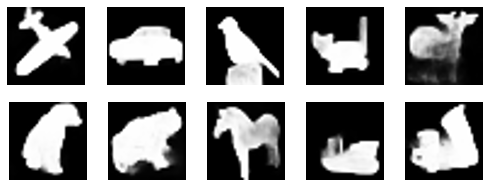}
   \caption{Examples of the manually selected silhouette images from CIFAR-10.}
   \label{fig: silhouette}
\end{figure}

To prove our second assumption, contrastive features contain less shape information than supervised features, we created a silhouette dataset based on CIFAR-10. As CIFAR-10 does not have a public silhouette dataset, We use a pretrained semantic segmentation model U2-Net \cite{Qin_2020_PR} to generate silhouette images. Due to the low resolution and no fine tuning, some generated images are of low quality. We manually selected 261 high-quality silhouette images to ensure that human beings are capable of making the right predictions only by their contour. Figure \ref{fig: silhouette} shows some examples. Table \ref{tab: shape biased} shows that the supervised model achieves the highest accuracy among all models. Such accuracy results to some extent could show that supervised features have more shape information than contrastive features.

\begin{table}
  \centering
  \begin{tabular}{c c c c c}
    \toprule
    Model & MoCo & SimCLR & SimSiam & Supervised\\ 
    \midrule
    Acc & 55.17\% & 46.7\% & 56.7\% & 60.91\% \\
    \bottomrule
  \end{tabular}
  \caption{Accuracy between different models on silhouette CIFAR-10 dataset.}
  \label{tab: shape biased}
\end{table}

\subsection{Distorted Image Evaluation}
\label{sec: Distortion}
The results of the above two experiments raise questions about the robustness of contrastive representations. Therefore, we did different distortions on the test set and reevaluate the accuracy. More specifically, we apply "ColorJitter", "HorizontalFilp", and "RandomRotation" respectively to the testset. Table \ref{tab: distortion} shows that all contrastive models have a lower decrease of accuracy in "ColorJitter" and "GrayScale" than supervised models, and there is no obvious difference in "Flip". The results indicate that contrastive representation is more robust than supervised representation for the knowledge that it has gained.

\begin{table}
  \centering
  \begin{tabular}{c c c c c c}
    \toprule
    model & baseline & color & flip & gray\\ 
    \midrule
    MoCo & 88.14\% & -1.22\% & +0.16\% & -3.35\% \\  
    SimCLR & 87.78\% & -1.15\% & -0.42\% & -4.40\% \\ 
    SimSiam & 88.43\% & -0.43\% & +0.10\% & -3.63\% \\
    \bottomrule
    Supervised & 95.18\% & -2.2\% & +0.22\% & -5.39\% \\
    \bottomrule
  \end{tabular}
  \caption{Accuracy change amplitude of each model for distorted images compared to the original baseline. All contrastive models have lower decrease of accuracy in "ColorJitter" and "GrayScale" than supervised models, and there is no obvious difference in "Flip".}
  \label{tab: distortion}
\end{table}

\section{Discussion}
As noted in Section \ref{sec:intro}, researchers have been working hard to improve unsupervised learning, and the quantitative improvement is usually compared with the accuracy of supervised learning. To discover fine-grained gaps between unsupervised and supervised learning, we proposed GAP-265, a subset of CIFAR-10 which unsupervised models make the same wrong predictions while the supervised model makes the right predictions. Most of these wrong predictions are due to the color of the object. Section \ref{sec: color biased} illustrate that contrastive representation space is more dependent on color than supervised representation space. We also constructed a silhouette dataset based on CIFAR-10 which contains 261 high-quality images. Section \ref{sec: Shape biased} illustrate that supervised model are more shape-aware than unsupervised models. Section \ref{sec: Distortion} illustrate that unsupervised models are more robust than the supervised model.

In our opinion, the reason why such phenomena appear is because of the data augmentation. Unsupervised models inevitably have to apply stronger augmentations such as "ColorJitter" and "GrayScale" than supervised models due to the annotation limitation. As a result, for these distorted images, unsupervised models are more robust than supervised models because that's how they were trained. However, this also leads to the problem of possible color dependence of the representation spaces trained by contrastive learning. By contrast, current augmentation methods do not have such a strong transformation of image shape as "ColorJitter" transforms image color. The more difficult thing is that it is unsupervised learning. We can change the color distribution of the image by adjusting the contrast and brightness. However, the shape distribution of the image is difficult to make changes on a 2D image without relying on labels. For example, it is easy to change the color of the object in one image, but it is obviously impossible to obtain images from all angles of the object through such one image.

\section{Conclusion}
In summary, we provided evidence that representation space based on contrastive learning today relies on color information more and contains less shape information compared with space based on supervised learning. In addition, we also indicate that contrastive space are more robust than supervised space on distortion attacks. We discussed that data augmentation is the cause of these problems. We hope that these experiments and conclusions can make researchers focus on how to add shape knowledge in the future unsupervised learning and add color as well as shape metrics to the evaluation metrics of unsupervised tasks.

\bibliographystyle{unsrt}  
\bibliography{references}

\end{document}